\documentclass{article}

 \PassOptionsToPackage{numbers, compress}{natbib}
\usepackage[final]{nips_2016}

\usepackage[utf8]{inputenc} 
\usepackage[T1]{fontenc}    
\usepackage{hyperref}       
\usepackage{url}            
\usepackage{booktabs}       
\usepackage{amsfonts}       
\usepackage{nicefrac}       
\usepackage{microtype}      
\usepackage[inline]{enumitem}

\usepackage{tikz}
\usetikzlibrary{automata,arrows,positioning,calc}

\usepackage{graphicx}
\usepackage{caption}
\usepackage{subcaption}

\usepackage{bm}
\usepackage{xcolor}
\usepackage{xspace}
\usepackage{amsmath}

\newcommand*{\ie}{i.e.\@\xspace}

\newcommand*{\RR}{{\sl RR}\@\xspace}
\newcommand*{\SP}{{\sl SP}\@\xspace}
\newcommand*{\PP}{{\sl PP}\@\xspace}
\newcommand*{\PROMs}{{\sl PROMs}\@\xspace}

\newcommand{\pd}{\mbox{{\sc palladio}}\@\xspace}
\newcommand{\mn}{\mbox{{\sc minimal}}\@\xspace}

\newcommand{\norm}[1]{\left\lVert#1\right\rVert}

\title{A temporal model for multiple sclerosis course evolution}

\author{
Samuele~Fiorini$^\dagger$,  Andrea~Tacchino$^\ddagger$, Giampaolo~Brichetto$^\ddagger$, Alessandro~Verri$^\dagger$,     Annalisa~Barla$^\dagger$\\
  Universit\`a degli Studi di Genova (DIBRIS) -
  Via Dodecaneso 35, Genova, Italy\\
   $^\ddagger$Italian Multiple Sclerosis Foundation, Scientific Research Area -
  Via Operai 40, Genova, Italy \\
     \texttt{samuele.fiorini@dibris.unige.it},
    \texttt{\{alessandro.verri, annalisa.barla\}@unige.it}, \\
    \texttt{\{andrea.tacchino, giampaolo.brichetto\}@aism.it} \\
}

\begin{document}

\maketitle

\begin{abstract}
Multiple Sclerosis is a degenerative condition of the central nervous system that affects nearly 2.5 million of individuals in terms of their physical, cognitive, psychological and social capabilities. Researchers are currently investigating on the use of patient reported outcome measures for the assessment of impact and evolution of the disease on the life of the patients. To date, a clear understanding on the use of such measures to predict the evolution of the disease is still lacking.
In this work we resort to regularized machine learning methods for binary classification and multiple output regression. We propose a pipeline that can be used to predict the disease progression from patient reported measures. The obtained model is tested on a data set collected from an ongoing clinical research project.


\end{abstract}

\section{Introduction}\label{sec:intro}
In the recent past, researchers explored the potential role of Patient Reported Outcome Measures (\PROMs), consisting in ordinal-scaled questionnaires and self-reported measures, to follow the progression of neurodegenerative diseases and  to take timely healthcare decisions.
Such patient-friendly and low-cost measures allow to investigate on changes and individual impact of disease
on many different domains of the patient's life including: physical, cognitive, psychological, social functioning and general well-being \citep{fiorini2015machine}.
Nevertheless, for many diseases there is still no evidence of which are the most informative \PROMs and, contextually, of whether it is possible to make predictions on the disease evolution based on information extracted from them.

Multiple Sclerosis (MS) is a neurodegenerative and chronic disease of the central nervous system characterized by damages to the myelin sheaths. MS patients experience a wide range of disorders, such as: fatigue, numbness, visual disturbances, bladder problems and mobility issues. MS patients are typically grouped in three classes according to their disease course:
\begin{enumerate*}[label=\alph*)]
  \item relapsing-remitting (\RR),
  \item secondary-progressive (\SP) and
  \item primary-progressive (\PP)
\end{enumerate*} \citep{giovannoni2016brain}.
About $85\%$ of people with MS have \RR form of the disease. This form is characterized by clearly defined {\sl relapses}, \ie  attacks of neurological worsening, followed by partial or complete recovery.
If left untreated, about $60\%$ of \RR patients develop \SP form within $15\text{--}20$ years and it takes only $14$ years on average for people to become unable to walk for $100$ meters unaided \citep{scalfari2014onset}. Patients in \SP form experience a steady progress of the disease. Finally, about $15\%$ of people with MS have a \PP form from the onset, with a gradual  progression of clinical symptoms in the absence of relapses.

In this context, the identification of the transition point when \RR converts to \SP is one of the most important methodological gaps that researchers are currently addressing. Nowadays, there are no clear clinical, imaging, immunologic or pathologic criteria to foresee the transition from \RR~to~\SP~\citep{lublin2014defining, vukusic2003prognostic}.

In this work, we propose a \PROMs-based data analysis pipeline that aims at filling this scientific gap.
In the context of neurodegenerative diseases, clinicians are used to take advantage of \PROMs data collections to corroborate evidences coming from standard quantitative exams \citep{black2013patient}. Interestingly, in our case the absence of clear \SP predictors makes the information extracted from \PROMs data the only available resource.

In our study, we investigate on the use of machine learning methods for the prediction of MS disease course over time, focusing on \RR to \SP transition. In particular, we take advantage of regularization methods, that is a class of techniques widely used in the biomedical context as they benefit from good generalization properties as well as they allow to solve both regression and classification problems within the same statistical and computational framework~\citep{azencott2013efficient, nowak2011fused, zou2005regularization}.

The proposed pipeline is tested on a data set
collected from MS patients currently enrolled in an ongoing funded project.

\section{Background}


Regularization methods are a popular class of machine learning techniques that can be expressed as the minimization problem in Equation~\eqref{eq:losspen}, where $X \in \mathbb{R}^{n \times d}$ and $Y \in \mathbb{R}^{n \times k}$ are matrices storing input data and output labels, respectively, for arbitrary $k$.

\begin{equation}\label{eq:losspen}
	\min_f V(f(X),Y) + \lambda R(f)
\end{equation}

The loss function $V(\cdot,\cdot)$ can be seen as a measure of {\sl adherence} to the training data; instead, the regularization penalty $R(\cdot)$ introduces additional information that is used to solve the problem.
The regularization parameter $\lambda$ controls the trade-off between the two terms.
In this work we focus on linear models, \ie $f = X \cdot W$, where $W \in \mathbb{R}^{d \times k}$.
In this case, the regularization penalty $R(\cdot)$ is often expressed as a norm of $W$. For scalar outputs (\ie~when $k=1$), combinations of $\ell_p$-norms are commonly used, while in case of multiple outputs (\ie~when $k>1$), $R(\cdot)$ is generally an $L_{p,q}$- or a {\sl Schatten} $p$-norm.
With different choices for $R(\cdot)$, different effects on the solution may be achieved.
We refer to \citep{gramfort2012mixed} for a thorough description of the most common regularization penalties.
For the sake of simplicity, we restrict to the case where $V(\cdot,\cdot)$ is a convex differentiable loss function and the regularization penalty $R(\cdot)$ is a non-differentiable term~\citep{hastie2015statistical}.

Proximal gradient methods \cite{combettes2005signal} are a popular class of optimization algorithms that
can be used to achieve the solution of a minimization problem of the form expressed in Equation~\eqref{eq:losspen}.
They can be considered as a generalization of classical gradient methods to non-differentiable convex optimization problems. These methods concatenate a gradient descent step with a non-linear {\em proximity} operation that can be seen as a generalized form of projection \cite{combettes2005signal}.
When learning a linear model, under few hypotheses on $V(\cdot,\cdot)$ and $R(\cdot)$ \cite{combettes2005signal}, the $j$-th proximal gradient step can be written as in Equation~\eqref{eq:proxgrad}.

\begin{equation}\label{eq:proxgrad}
W^{j} := \mathrm{prox}_{\lambda R(\cdot)}(W^{j-1} - \gamma\nabla V(X\cdot W^{j-1},Y))
\end{equation}

According to $R(\cdot)$, the proximal mapping of Equation~\eqref{eq:proxgrad} is expressed in different forms. See \citep{gramfort2012mixed} for a comprehensive description of the proximal mapping of the most common regularization penalties.


\section{Proposed pipeline}
We consider $T$ sets of $n$ input-output pairs $\{\bm{x}^{t}_i, y^{t}_i\}_{i=1}^n$ for $t=1,\dots,T$, where $\bm{x}_i^{t}\in\mathbb{R}^d$ is a $d$-dimensional representation of the $i$-th MS patient at time point $t$.
In this paper the representation of each patient at a fixed time point consists in a $d$-dimensional vector carrying the ordinal-scaled answers to the set of \PROMs.
In this work we focus on the \RR to \SP transition considering the latter as the positive class, so $y_i^{t} \in \{\pm 1 \}$ is a binary label corresponding to the MS disease course diagnosed for the $i$-th patient at time point $t$.

The proposed pipeline aims at solving two related problems, namely {\sl diagnosis} and {\sl prognosis}.
The diagnosis problem, given a $d$-dimensional representation of a patient at a fixed time point $\bm{x}_i^t$, consists in detecting the corresponding disease course $y_i^t$.
Given a patient and the corresponding historical representation $\bm{x}_i^t$ for $t=1,\dots,T$, the prognosis problem consists in predicting if the patient at $t=T+1$ will experience a transition from \RR to \SP or remain at the same pathological stage.

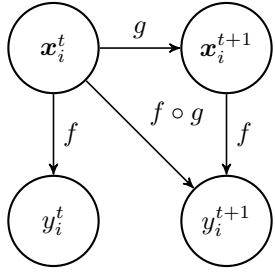
\begin{figure}
  \centering
  \begin{tikzpicture}[->, >=stealth', auto, semithick, node distance=2.3cm]
    \tikzstyle{every state}=[fill=white,draw=black,thick,text=black,scale=1, minimum height=1.2cm]
    \node[state]    (xit)                     {$\bm{x}_i^t$};
    \node[state]    (yit)[below of=xit]       {$y_i^t$};
    \node[state]    (xit1)[right of=xit]      {$\bm{x}_i^{t+1}$};
    \node[state]    (yit1)[below of=xit1]     {$y_i^{t+1}$};
    \path
    (xit)  edge node{$f$} (yit)
    (xit)  edge node{$g$} (xit1)
    (xit1) edge node{$f$} (yit1)
    (xit)  edge node{$f\circ g$} (yit1);
  \end{tikzpicture}
  \caption{A visual representation of the temporal structure assumed in the input-output pairs. Learning the two functions $f$ (diagnosis) and $g$ (\PROMs time evolution) it is possible to predict the trend of the disease course $y_i^t$ for future time points with  $f\circ g$ (prognosis).}\label{fig:pipeline}
\end{figure}

The proposed pipeline aims at learning a statistical model to solve diagnosis and prognosis problems assuming the temporal structure of input-output pairs of Figure~\ref{fig:pipeline}.
First, the diagnosis problem
can be modeled as $f(\bm{x}_i^{t}) = y_i^{t}$.
Secondly, the evolution over time of the input data can be written as $g(\bm{x}_i^{t}) = \bm{x}_i^{t+1}$. Once $f(\bm{x})$ and $g(\bm{x})$ are learned by training on historical data, the prognosis problem can be modeled by considering $f \circ g(\bm{x}_i^{t}) = y_i^{t+1}$.

Our experimental design consists in training $f(\bm{x})$ and $g(\bm{x})$ on the data collected at time points $t = 1, \dots ,T'$ ({\sl learning set}) and test them on the remaining data at $t = T'+1, \dots, T$ ({\sl test set}).
The two minimization problems for  learning $f(\bm{x})$ and $g(\bm{x})$ are solved following the {\sl Fast Iterative Shrinkage-Thresholding Algorithm} \citep{Beck09}.
The proposed pipeline consists of the following three steps.

\begin{figure}[t]
    \centering
    \begin{subfigure}[b]{0.4\textwidth}
        \includegraphics[width=\textwidth]{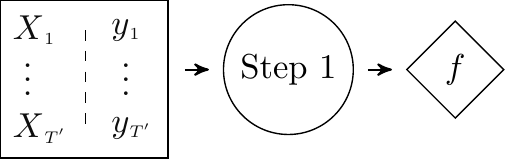}
        \caption{}
        \label{fig:step1}
    \end{subfigure}
    \hfill
    ~ 
    \begin{subfigure}[b]{0.4\textwidth}
        \includegraphics[width=\textwidth]{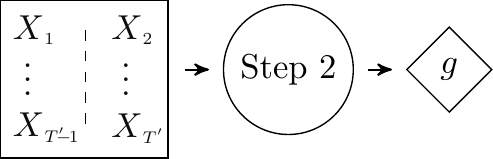}
        \caption{}
        \label{fig:step2}
    \end{subfigure}
    \caption{A visual representation of the learning steps for $f$ (diagnosis) and $g$ (\PROMs time evolution) acting on the learning sets. The vertical dashed lines within the rectangles separate input, on the left, and output, on the right, of the learning step.}
    \label{fig:steps}
\end{figure}

\begin{figure}[b]
  \centering
  \includegraphics[width=0.5\textwidth]{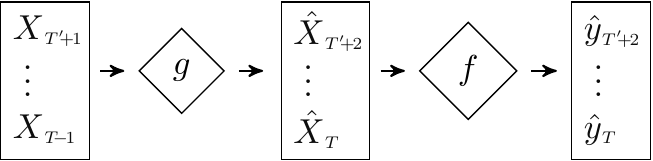}
  \caption{A visual representation of the Step 3. The prognosis model $f \circ g$, trained on historical data, predicts the evolution of the disease course of the patients for future time points (test set).}
  \label{fig:fog}
\end{figure}

\begin{description}
  \item[Step 1]
  With respect to the notation introduced in Equation~\eqref{eq:losspen},
  the single output classifier $f(\bm{x})$ is learned minimizing an objective function that combines the square loss \mbox{$V(\cdot,\cdot) = \frac{1}{N} \norm{X \cdot \bm{w} - \bm{y}}_2^2$} and the \mbox{$\ell_1\ell_2$-penalty} \mbox{$R(\cdot)=\mu\norm{\bm{w}}_2^2 + \tau\norm{\bm{w}}_1$}.
  The data matrix $X \in \mathbb{R}^{N\times d}$ and the output vector $\bm{y} \in \{\pm1\}^N$ carry, respectively, \PROMs and disease course of the patients at time points $t=1,\dots,T'$,  as in Figure~\ref{fig:step1}. $N$ is the total number of learning samples.
  The combined influence of the $\ell_1$- and the $\ell_2$-norm of $R(\cdot)$ achieves a {\sl sparse} but {\sl stable} solution in which grouped selection of collinear variables is promoted \citep{de2009regularized}. Therefore, this learning step attains a variable selection effect as it identifies
  a weight vector $\bm{w}$ with only $\widetilde{d} < d$ non-zero entries corresponding to the meaningful variables.
  The statistical robustness of the obtained results, in terms of prediction and variable selection, is guaranteed by a Monte Carlo resampling scheme implemented in the open-source project \pd~\citep{barbieri16palladio}.

  \item[Step 2]
  The time evolution of \PROMs, modeled by the multiple output function $g(\bm{x})$, is learned leveraging on a vector-valued regression scheme.
  According to Figure~\ref{fig:step2}, data matrices $X \in \mathbb{R}^{N'\times d}$ and $Y \in \mathbb{R}^{N'\times \widetilde{d}}$ contain the ordinal answers provided by $N'$ patients at time points $t=1,\dots,T'-1$ and $t=2,\dots,T'$, respectively.
  By design, the function $g(\bm{x})$ predicts only the evolution of the $\widetilde{d}$ meaningful variables selected at Step 1.
  The loss function of choice is the squared {\sl Frobenius} norm $V(\cdot,\cdot) = \frac{1}{N'} \norm{X \cdot W - Y}_F^2$ while the chosen regularization penalty  $R(\cdot)$ is the mixed $L_{2,1}$-norm, \ie~the sum of the $\ell_2$-norm of the rows of the weight matrix $W\in\mathbb{R}^{d\times\widetilde{d}}$.
  This choice of $R(\cdot)$ promotes a row-structured sparsity pattern in the solution \citep{gramfort2012mixed, evgeniou2007multi}.
  $g(\bm{x})$ is learned taking advantage of the implementation offered by the open-source project \mn\footnote{\url{https://github.com/samuelefiorini/minimal.git}}.


  \item[Step 3]
  As the two models $f(\bm{x})$ and $g(\bm{x})$ are trained on the learning set, a reliable estimation of the predictive power of $f \circ g(\bm{x})$ can be evaluated on the independent test set. As in Figure~\ref{fig:fog}, the disease course evolution $f \circ g(\bm{x}^{t}) = \hat{y}^{t+1}$ is evaluated for $t = T'+1,\cdots,T-1$.
  The predictive power of the final model is finally assessed measuring the rate of agreement of $\hat{y}^t$ with respect to $y^t$.

\end{description}

  Prognosis, in daily clinical practice, can be achieved for future time points running predictions $f\circ g(\bm{x}^t)$ for $t=T,\dots,T+\Delta T$, where $\Delta T$ is an arbitrarily long prediction horizon.


\section{Experiments and results}\label{sec:experiments}

In this section, we present the results obtained by running our pipeline on a set of \PROMs collected every four months from a cohort of MS patients enrolled in the ongoing study {\sl DETECT-MS PRO} \citep{fiorini2015machine}. At the time of the analysis, the data collection was lasting for two years, which means we had data for $t=1,\dots,T$ with $T=6$ time points. Learning and test sets consisted of the first $t=1,\dots,T'$ with $T'=4$, and the last two time points ($t=5,6$), respectively.
The dimensionality of each sample is $d=145$ corresponding to the number of submitted \PROMs.
As patients enrollment was still ongoing, the number of individuals reaching the $t$-th examination $n_t$ decreases for increasing $t$. According to our experimental design, for the diagnosis problem in Figure~\ref{fig:step1} we had $N=\sum_{t\leq T'}n_t=2023$ learning and $N_{test}=\sum_{t>T'}n_t=194$ test samples.
The learning process of $f(\bm{x})$ led to the identification of $\widetilde{d}=16$ relevant variables. Considering $100$ Monte Carlo resampling the corresponding average balanced classification accuracy~\citep{brodersen2010balanced} was $80.0\%~(\pm1.0\%)$.

Given the structure of the vector-valued regression problem of Step 2 as in Figure~\ref{fig:step2}, the cardinality of the learning set was $N'=\sum_{t\leq T'-1}n_t=1292$ while the test set consisted of $N'_{test}=\sum_{t>T'+1}n_t=62$ samples.
The multiple output regression model $g(\bm{x})$ identified on the learning set was used to predict the time evolution of the \PROMs of the test set $g(\bm{x}_i^{t-1})=\hat{\bm{x}}_i^t$ for $t=T'+1,\dots,T-1$.

Finally, we solved the prognosis problem of Figure~\ref{fig:fog} for the time points $t=T'+2,\dots,T$ estimating $f \circ g(\bm{x}_i^{t-1})=f(\hat{\bm{x}}_i^t)=\hat{y}_i^t$ for $t=T'+1,\dots,T-1$. We compared the predictions $\hat{y}_i^t$ obtained by our model $f \circ g(\bm{x})$ with the courses $y_i^t$ assigned by clinicians. The concordance rate reached $80.3\%$.

\section{Conclusion}\label{sec:conclusion}
In this work we presented a data analysis pipeline based on regularized machine learning methods that addresses diagnosis and prognosis problems in the context of MS. The proposed pipeline was extensively validated on synthetic problems (results not shown) and it was tested on a real-world data collection. This set of data consisted of a large number of ordinal answers provided by a cohort of individuals to a given set of \PROMs. Our statistical model was able to correctly predict the evolution of the disease form of $80.3\%$ of the patients at the last time point.
Moreover, our result shows that a timely prediction of the disease course can be obtained from patient-friendly and low-cost \PROMs.
To the best of our knowledge, this is the first attempt of defining a quantitative decision support system for this task.
Finally, we remark that clinicians involved in this funded project will soon start the validation of the system in their daily practice.



\subsubsection*{Acknowledgments}
Supported by FISM - Fondazione Italiana Sclerosi Multipla - cod. 2015/R/03. The authors also gratefully acknowledge the support of NVIDIA
Corporation with the donation of the Tesla K40c GPU used
for this research.


\end{document}